\documentclass{ecai} 
\usepackage{microtype}
\usepackage{graphicx}
\usepackage{subfigure}
\usepackage{booktabs} % for professional tables

\usepackage{algorithm}
\usepackage{algorithmicx}
\usepackage{algpseudocode} % Dla bardziej zaawansowanych funkcji

\usepackage{amsmath}
\usepackage{amssymb}
\usepackage{mathtools}
\usepackage{amsthm}
\usepackage{array, makecell}
\usepackage{wrapfig}
\usepackage{multirow}

\usepackage{hyperref}

\usepackage[capitalize,noabbrev]{cleveref}

\usepackage[textsize=tiny]{todonotes}

\usepackage{cuted}
\usepackage{capt-of}
\theoremstyle{plain}
\newtheorem{theorem}{Theorem}[section]
\newtheorem{proposition}[theorem]{Proposition}

\theoremstyle{definition}

\theoremstyle{remark}

\def\exit{\mathrm{Exit\_Time}}
\def\NE{\mathrm{NET}}

\def\our#1{PrAViC#1}

\begin{document}

%%%%%%%%%%%%%%%%%%%%%%%%%%%%%%%%%%%%%%%%%%%%%%%%%%%%%%%%%%%%%%%%%%%%%%%%

\begin{frontmatter}

%%% Use this command to specify your submission number.
%%% In doubleblind mode, it will be printed on the first page.

\paperid{3764} 

%%% Use this command to specify the title of your paper.

\title{PrAViC: Probabilistic Adaptation Framework for Real-Time Video Classification}

%%% Use this combinations of commands to specify all authors of your 
%%% paper. Use \fnms{} and \snm{} to indicate everyone's first names 
%%% and surname. This will help the publisher with indexing the 
%%% proceedings. Please use a reasonable approximation in case your 
%%% name does not neatly split into "first names" and "surname".
%%% Specifying your ORCID digital identifier is optional. 
%%% Use the \thanks{} command to indicate one or more corresponding 
%%% authors and their email address(es). If so desired, you can specify
%%% author contributions using the \footnote{} command.
\author[ACD]{\fnms{Magdalena}~\snm{Tr{\k{e}}dowicz}}
\author[A]{\fnms{Marcin}~\snm{Mazur}}%\orcid{0000-0002-3440-8173}}
\author[ACD]{\fnms{Szymon}~\snm{Janusz}}
\author[BC]{\fnms{Arkadiusz}~\snm{Lewicki}}%\orcid{0000-0002-1922-1562}}
\author[AC]{\fnms{Jacek}~\snm{Tabor}}%\orcid{0000-0001-6652-7727}}
\author[AC]{\fnms{\L{}ukasz}~\snm{Struski}%\orcid{0000-0003-4006-356X}
\thanks{Corresponding Author. Email: lukasz.struski@uj.edu.pl.}}

\address[A]{Faculty of Mathematics and Computer Science, Jagiellonian University}
\address[B]{Faculty of Applied Computer Science, University of Information Technology and Management in Rzeszow}
\address[C]{Prometheus MedTech.AI}
\address[D]{Doctoral School of Exact and Natural Sciences, Jagiellonian University}

%%% Use this environment to include an abstract of your paper.

\begin{abstract}
Video processing is generally divided into two main categories: processing of the entire video, which typically yields optimal classification outcomes, and real-time processing, where the objective is to make a decision as promptly as possible. Although the models dedicated to the processing of entire videos are typically well-defined and clearly presented in the literature, this is not the case for online processing, where a~plethora of hand-devised methods exist. To address this issue, we present \our{}, a novel, unified, and theoretically-based adaptation framework for tackling the online classification problem in video data. The initial phase of our study is to establish a mathematical background for the classification of sequential data, with the potential to make a decision at an early stage. This allows us to construct a natural function that encourages the model to return a result much faster. The subsequent phase is to present a straightforward and readily implementable method for adapting offline models to the online setting using recurrent operations. Finally, \our{} is evaluated by comparing it with existing state-of-the-art offline and online models and datasets. This enables the network to significantly reduce the time required to reach classification decisions while maintaining, or even  enhancing, accuracy.
\end{abstract}

\end{frontmatter}

%%%%%%%%%%%%%%%%%%%%%%%%%%%%%%%%%%%%%%%%%%%%%%%%%%%%%%%%%%%%%%%%%%%%%%%%

\section{Introduction}

In recent years, there has been a notable increase in the utilization of convolutional neural networks (CNNs) across various fields where the capacity to make expeditious decisions could be crucial. This includes areas such as medicine~\citep{krenzer2023real,sapitri2023deep}, human action recognition (HAR) ~\citep{mollahosseini2016going,yang2023yowov2}, and autonomous driving~\citep{wu2017squeezedet}. However, despite the growing prevalence of CNNs in these domains, there remains a lack of a unified approach to the problem of making early decisions based solely on the initial frames.

\begin{figure}[ht!]
    % \vspace{-15pt}
    \centering
    \includegraphics[width=\linewidth]{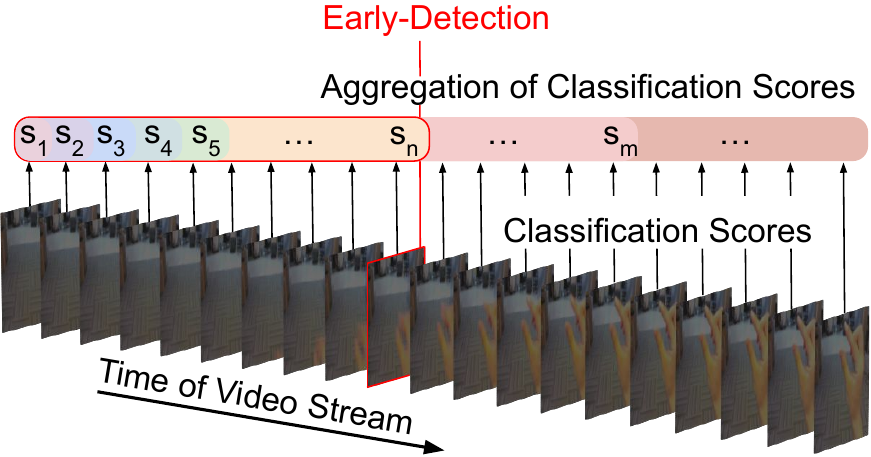}
    \caption{Our approach consists of three components: an online classifier generating Classification Scores, an Aggregation module, and an Early-Detection mechanism.}
    \label{fig:teaser_PrAViC}
    \vspace{15pt}
\end{figure}

Conversely, numerous offline approaches have been proposed (see, e.g., ~\citep{bhola2024review,karim2024human,kaseris2024comprehensive,ming2024action,yao2019review}) to address the problem of video data classification. However, these models usually require access to entire videos, which precludes their real-time applicability. Although certain techniques have been developed to facilitate the adaptation of offline models to the online domain (e.g., those proposed by \cite{kopuklu2022dissected,xiao2023realtime}), there remains a need to develop novel solutions that can deliver high performance across different types of data.

% In contrast to traditional methodologies, our approach allows for the adoption of existing 3D CNN models, wherein subtle adjustments are made to leverage the strengths of traditional CNN-3D networks while addressing specific challenges related to depth processing and feature extraction. 
%%%%%%%%%%%%%%%%%%%%%%%%%%
To address this gap we propose a novel \textbf{Pr}obabilistic \textbf{A}daptation Framework for Real-time \textbf{Vi}deo \textbf{C}lassification (\our{}, see Fig.~\ref{fig:teaser_PrAViC}). This framework provides a simple yet effective method to transform offline models for online use. Additionally, an early exit mechanism in the loss function is incorporated into \our{}, ensuring that the model is trained to deliver fast and accurate predictions suitable for dynamic, real-time environments.  It is important to note that while early exit strategies play an important role in our approach, the primary focus was on integrating these mechanisms with online methods to create a seamless and efficient system for video data processing. The ability to adapt offline, pre-trained 3D CNN models for real-world applications hinges on enabling them to make timely decisions.
%%%%%%%%%%%%%%%%%%%%%%%%%%%%

In addition, the \our{} strategy facilitates recursive utilization, a consequence of the novel method we propose for processing the network's head. The implications of such a development are far-reaching, impacting a multitude of domains including industry, medicine, and public safety. In these fields, the capacity to conduct rapid real-time analysis is of paramount importance for informed decision-making and the implementation of proactive measures.

In the experimental study, we demonstrate the efficacy of our approach when applied to a selection of state-of-the-art offline and online models trained on three real-world datasets, including four publicly available datasets, UCF101~\citep{soomro2012ucf101}, EgoGesture~\citep{zhang2018egogesture}, Jester~\citep{materzynska2019jester}, and Kinetics-400~\citep{kay2017kinetics}, as well as a closed-access real ultrasound dataset, comprising Doppler ultrasound images representing short-axis and suprasternal views of newborn hearts.
% Moreover, we introduce an innovative function that enables the model to make earlier exits (decisions) when sufficient evidence is accumulated. 
Moreover, the incorporation of the proposed early exit mechanism appears to have been evidenced to optimize the timing of \our{}'s decisions, thereby enhancing efficiency, albeit with occasional slight loss of accuracy.

In conclusion, our contributions can be summarized as follows:
\begin{itemize}
    \item we introduce \our{}, a novel, unified, and the\-or\-etically-based probabilistic adaptation framework for online classification of video data, which encourages the network to make a decision at an early stage (see Section~\ref{sec:pravic}),
    \item we propose a straightforward and easily implementable method to adapt offline video classification models with 3D CNN architectures for online use, based on a~ specific customization of the head design, ready to be integrated with the \our{} approach (see Section~\ref{section:architecture}),
    \item we conduct experiments that demonstrate the effectiveness of \our{} and its ability to facilitate earlier classification decisions while maintaining or even improving accuracy compared to selected state-of-the-art offline and online solutions (see Section~\ref{sec:experiments}).
\end{itemize}
%%%%%%%%%%%%%%%%%%%%%%%%%%%%%%

\section{Related Works}

While video-based 3D networks have been widely studied in offline settings ~\citep{bhola2024review, karim2024human,kaseris2024comprehensive, ming2024action,yao2019review}, developing online models remains still a challenge. Such online models enable real-time classification, facilitating applications such as emergency situations detection or medical diagnostics.

Considering video as a series of consecutive frames, each frame may be classified individually using 2D CNN models \citep{naturaldieseases}. This approach represents an online technique and demonstrates effectiveness in real-time classification, due to the limited number of parameters involved. However, the absence of temporal information from the entire video may present a limitation, potentially leading to misclassifications such as predicting ``sitting down'' when the action involves transitioning from sitting to standing. Consequently, to develop a model capable of processing online data in real time, 2D networks are often utilized, complemented by additional mechanisms for managing temporal data~\citep{chang2024lm,shen2023real,wang2021action,xiao2023realtime,xu2023improving}. For instance, the authors of~\citep{xiao2023realtime} employed a temporal shift module (TSM). Their approach involves shifting a portion of the channels along the temporal dimension, thereby facilitating the capture of temporal relationships. 

% \paragraph{Online 3D models for real-time usage}
On the other hand, it has been demonstrated that 3D CNNs yield superior accuracy in video classification tasks compared to 2D CNNs~\citep{carreira2017quo}. Consequently, novel approaches have been devised that utilize distinct versions of 3D convolutional kernels. Several studies have proposed the development of dedicated architectures for online 3D networks, including those presented by~\cite{kim2024atrous,krenzer2023real,sapitri2023deep,yang2023yowov2}, which are designed to operate in real time. 
Another commonly utilized approach is the combination of 3D CNN architectures with various long short-term memory (LSTM) models. For example, \cite{lu2024real} introduce a DNN that merges 3D DenseNet variants and BiLSTM. In turn, \cite{chen2023parallel} propose a combination of R2plus1D and ConvLSTM in a parallel module. The proposed network utilizes the attention mechanism to extract the features that require attention in the channel and the spatial axes.

An alternative approach to the problem of online video classification, as proposed by~\cite{ustek2023two}, involves combining a vision transformer for human pose estimation with a CNN-BiLSTM network for spatio-temporal modelling within keypoint sequences. Similarly, attention mechanisms, in conjunction with transformer layers, have also been employed in~\citep{huang2023real}. However, it should be noted that the adaptations implemented by our framework do not extend to architectures such as transformers or CNN-LSTMs.

The concept of converting various well-known resource-efficient 2D CNNs into 3D CNNs, as proposed by~\cite{kopuklu2020online}, is the most closely related to our approach. In this context, \cite{kopuklu2022dissected} investigate the potential of adapting 3D CNNs (particularly the 3D ResNet family of models) for online video stream processing. Their approach involves the elimination of temporal downsampling and the utilization of a cache to store intermediate volumes of the architecture, which can then be accessed during inference. A comparable solution has also been proposed by \cite{hedegaard2022continual}, who advanced the concept of weight-compatible reformulation of 3D CNNs, designated as Continual 3D Convolutional Neural Networks (Co3D CNNs). Co3D CNNs facilitate the processing of videos in a~ frame-by-frame manner, utilizing existing 3D CNN weights, thereby obviating the necessity for further fine-tuning.

%%%%%%%%%%%%%%%%%%%%%%%%%%%%%%

\section{Probabilistic Adaptation Framework for Real-time Video Classification (\our{})}\label{sec:pravic}

This section presents the details of the proposed \our{} model. The problem of video classification is divided into two categories: offline and online. In the offline case, where the entire video is available, classification can be performed by processing the entire video. In the online case, where consecutive images are obtained, the objective is to make a decision using only a partial subset of the potentially available frames. To facilitate comprehension, the discussion will initially focus on binary classification and subsequently transition to multiclass scenarios.

\paragraph{Standard Offline Case}
We assume that we are given a pre-trained neural network $\phi$, which for a given video $V=[V_0,\ldots, V_n]$ (where $V_i$ denotes the $i$-th frame) returns the probability $\phi(V)$ that $V$ belongs to the positive class. Finally, the decision is based on the threshold $\tau\in (0,1)$ (where typically $\tau=1/2$), i.e.,
\begin{equation} \label{eq:0}
\begin{array}{c}
V \text{ has the positive class} \Longleftrightarrow \phi(V) \geq \tau.
\end{array}
\end{equation}
An important observation is that we can compute the probability $\phi(V)$ if we know what the model's decision was for each threshold $\tau$.

\begin{proposition} \label{pr:1}
We have
\begin{equation} \label{eq:1}
\begin{array}{c}
\phi(V)=\mathrm{Prob}(\phi(V) \leq \tau: \tau \sim \mathrm{unif}_{[0,1]}).  
\end{array}
\end{equation}
\end{proposition}
The above formula can be interpreted as selecting a random threshold value $\tau$ from the interval $[0,1]$ and calculating the probability of being below the threshold. This is useful as in the case of online models, we have the natural definition of the threshold, and consequently it will allow to deduce the probabilistic model behind.

The subsequent paragraph addresses the question of how the offline model can be applied to the online procedure.

\paragraph{Standard Early Exit Approach to Online Classification} In this paragraph, we describe how we typically deal with applying a pre-trained model $\phi$ to the task of online classification, where we want to allow the model to make the final classification decision without the need to process all the frames.

We assume that the frames arrive consecutively\footnote{In the case of some architectures we allow them to arrive in groups of typically two, four, or eight frames.}. Thereafter, given a pre-trained video classification network $\phi$, we choose a threshold $\tau\in (0,1)$ and proceed with the following simple algorithm:

\begin{algorithmic}[1]
\State Set $k \gets 0$
\Repeat
    \State Load frame $V_k$ 
    \State Compute $p_k \gets \phi([V_0, \ldots, V_k])$
    \If{$p_k \geq \tau$}
        \State \Return class 1 (positive) for $V = [V_0, \ldots, V_n]$
    \Else
        \State $k \gets k + 1$
    \EndIf
\Until{$k = n + 1$}
\State \Return class 0 (negative)
\end{algorithmic}

The objective of the aforementioned procedure is to allow making the decision of $V$ being in a positive class before loading all frames from $V$. It should be noted that if we operate in the offline mode, where we have access to all data, the above algorithm can be constrained to computing $p=\max(p_1,\ldots, p_n)$, and then determining that the class of $V$ is negative if $p<\tau$, or positive otherwise.

\paragraph{Probabilistic Model behind \our{}}
Note that once the above procedure is completed, we only know the decision that was made, but we lack the information necessary to calculate the probability of the given outcome. Without this information, it is not possible to use the Binary Cross-Entropy (BCE) loss and fine-tune the model. To address this issue, the underlying concept of our proposed solution is to reinterpret the above approach so that it returns the probability behind the given decision. Specifically, we apply the probabilistic concept from Proposition~\ref{pr:1},
% defining the probability that $V$ belongs to a positive class as
% \begin{equation}\label{eq:positiveclass}
% \begin{array}{c}
% \Phi(V)=\max\{\phi([V_0]),\ldots,\phi([V_0,\ldots,V_n])\}.
% \end{array}
% \end{equation}
%The above probabilistic approach 
which allows us to compute the soft probability of the decision, knowing that the decision was made for an arbitrary threshold. This gives us the following theorem.
\begin{theorem}
The correct probability that $V$ has the positive class in the online case is given by
\begin{equation}\label{eq:pV}
\begin{array}{c}
p(V)=\max(p_0,\ldots,p_n), \text{ where }p_i=\phi([V_0,\ldots,V_i]).
\end{array}
\end{equation}
\end{theorem}

This is of paramount importance insofar as our objective is for the model to make more timely decisions.% (see Fig.~\ref{fig:PrAViC_decision}).

\begin{figure}[thb]
    \centering
    \includegraphics[width=\linewidth]{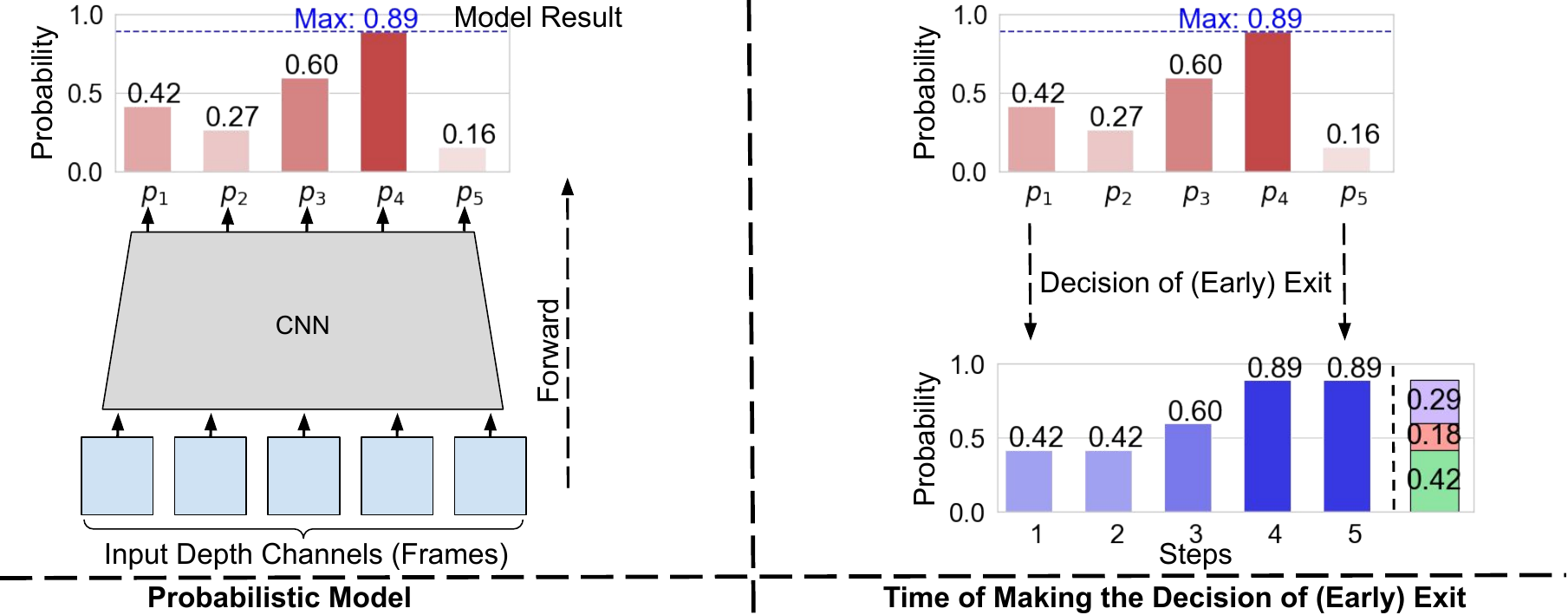}
    % \caption{The left image shows a probabilistic model where the outcome is determined by selecting the maximum value of $p_i$ (in this case, 0.89). The right image shows an alternative approach where the model can exit earlier based on the available probabilities. In this example, the model terminates with a probability of 0.42 at the 1st step, a probability of 0.18 at the 3rd step, and a probability of 0.29 at the 4th step.}
    \caption{Comparison of two probabilistic decision models. Left: A standard model selects the outcome with the highest probability (in this case, 0.89). Right: An alternative model allows for early exit based on intermediate probabilities. In this example, the process ends with a probability of 0.42 at step 1, 0.18 at step 3, and 0.29 at step 4.}
    \label{fig:PrAViC_decision}
    \vspace{15pt}
\end{figure}

%\paragraph{Expected time of making the decision of (early) exit} 
\paragraph{Expected Time of Early Exit}
To proceed to the probabilistic approach, we first calculate the expected time our model will make an early exit. We are interested in calculating the function $\exit(V)$, which represents the expected time of early exit for a given video $V$ (provided that it has been made).
% i.e.,
% \begin{equation}\label{eq:exit}
% \begin{array}{c}
% \exit(V)=\mathbb{E}(T_V|T_V<\infty).    
% \end{array}
% \end{equation}

\begin{theorem}
The expected exit time is given by
\begin{equation} \label{eq:t}
\exit(V)= n-\frac{1}{\max p_i}\sum\limits_{t=0}^{n-1}\max \limits_{i=0..t} p_i \in [0,n].
\end{equation}
\end{theorem}

\begin{proof}
Let $W$ be the random variable that returns the time at which we made an exit for a given video $V$, assuming that the model made an early exit at a time $k\in \{0,1,\ldots,n\}$. Consequently, we can sample from $W$ as follows: we draw a random variable $\tau$ uniformly from the interval $[0,\max_i p_i]$, and for the given value of $\tau$ we return the first index $k$ for which $p_k \geq \tau$.
Then we have
\begin{equation}\label{eq:exp}
\begin{array}{r@{\;}l}
\mathbb{E}(W) = & p(W\geq 1)+ \ldots + p(W\geq n) \\
= & n-P(W \leq 0)-\ldots -P(W \leq n-1),
\end{array}
\end{equation}
which gives the assertion.
\end{proof}

To make the function $\exit$ independent of the number of frames of $V$, we introduce its normalized version, given by
\begin{equation}\label{eq:exit}
\NE(V)=\tfrac{1}{n}\cdot \exit(V) \in [0,1].
\end{equation}
The function $\NE$ is defined as a normalization of $\exit$ with respect to the number of frames. It returns $0$ if we have made the exit (with probability one) on the frame $V_0$, and $1$ if we cannot exit before $V_n$ and the probability of exit in $V_n$ is positive.
\paragraph{Loss Function for \our{}}
Now we will describe how to incorporate the $\NE$ function into the objective in a way that encourages the model to exit earlier. However, we do not want to stimulate the model too much so as not to provoke wrong decisions. To do this, we set a parameter $\lambda \in [0,1]$ that tells us how large a percentage of confidence we are willing to potentially sacrifice in order to make the decision early. 
%Consider the videos $\bar V,\tilde V$ from the positive class, where $p(\bar V)=1,\NE(\bar V)=1$ and $p(\tilde V)=\tau,\NE(\tilde V)=0$. Our goal is to add an additional part to the loss function in such a way that the loss of $V$ and $\tilde V$ is equal. Obviously
%$\log((1-\lambda)+\lambda\NE(\bar V))=\log((1-\lambda)+\lambda\NE(\tilde V))$.
Accordingly, the value of \our{}'s objective for a frame $V$ with a class $y \in \{0,1\}$ is defined as follows:
\begin{equation}\label{eq:loss_function}
    \mathcal{L}_{\text{\our{}}_\lambda}(V,y)=\mathcal{L}_\text{BCE}(V,y)+y \log(\lambda+(1-\lambda)\NE(V)).
\end{equation}
It is evident that for videos belonging to the negative class (i.e., $y=0$), the loss function remains the standard BCE loss. 

% {\color{red}
% W przypadku wielu klas $y\in\mathrm{C}\subset\mathbb{N}$, funkcja kosztu jest dana wzorem:
% \begin{equation}
%     \mathrm{loss}_\lambda(V,y)=\mathrm{CE\_loss}(V,y) + \log(\lambda+(1-\lambda)\NE(V_{pred})).
% \end{equation}
% gdzie $\NE(V_{pred})$ obliczamy jak dla przypadku binarnego tylko zawężamy się do wartości $p_i$ na klasie wnioskowanej przez model (biorę maksymalne wartości logitu na wymiarze klas). 
% }

For $\lambda = 1$, the additional part of the loss function is equal to 0, which discourages the model from making early decisions. As the value of $\lambda$ approaches $0$, the model is encouraged to make decisions as rapidly as possible, even if this results in a loss of accuracy. Note that we mark the $\lambda$ for which the model was trained by using it as the subscript in \our{}$_\lambda$.

We emphasize that in the loss for the binary classification case given by  Eq.~\eqref{eq:loss_function}, we do not penalize the points with negative class, in other words we do not encourage the model to make earlier decisions in this case. This follows from our motivation coming from real life situations, where class $1$ corresponds to an emergency-type event (heart or machine failure, car accident, etc.) and should be detected as early as possible, while class $0$ corresponds to the default (normal) state of the system.

\paragraph{Loss Adjustment for Multiclass Case}
In the case of multiple classes, the value of \our{}'s objective for a frame $V$ with a class  $y\in \{1,\ldots,n\}$ can be defined as follows:
\begin{equation}\label{eq:loss_function_mulit}
\mathcal{L}_{\text{\our{}}_\lambda}(V, y) = \mathcal{L}_\mathrm{CE}(V, y) + \log(\lambda + (1 - \lambda) \NE(V_{\text{pred}})),
\end{equation}
where $\mathcal{L}_\mathrm{CE}$ denotes the standard Cross-Entropy loss and $\NE(V_{\text{pred}})$ is computed similarly to the binary case, but restricted to the value $p_i$ corresponding to the class predicted by the model (i.e., the class with the highest logit value along the class dimension). Note that in this case there is no negative class we need to penalize, rather we treat all classes as equally important to detect.
%%%%%%%%%%%%%%%%%%%%%%%%%%%%%%

\section{Architecture of the Model}
\label{section:architecture}

This section presents the modifications and extensions to the CNN-3D architecture that form the basis of our \our{} approach. They include specific changes to key layers, including the convolution and batch normalization layers, as well as a novel method for processing the network's head, which constitutes one of the primary findings of our work. These modifications are designed to leverage the strengths of traditional CNNs while addressing specific challenges related to depth processing and feature extraction. The following paragraphs provide a detailed breakdown of each component and its role in our approach.

\begin{figure}[thb]
    \centering
    \includegraphics[width=\linewidth]{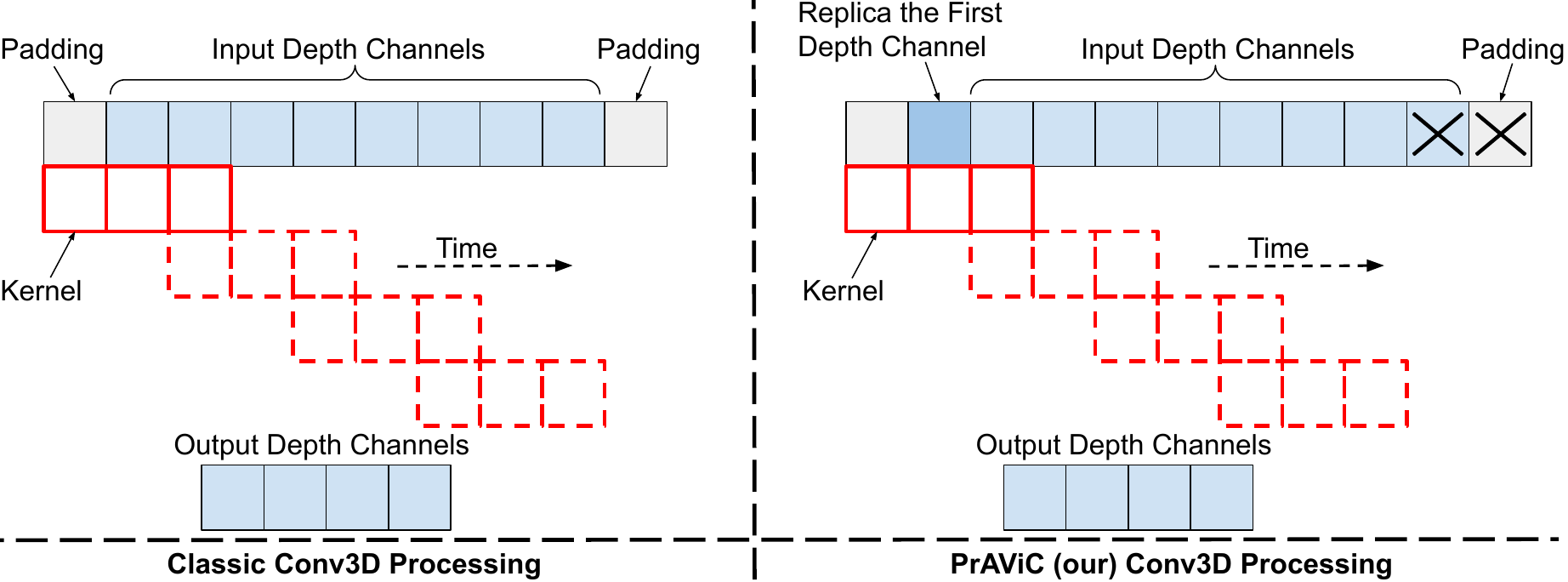}
    \caption{Illustration of the mechanism of the classical 3D convolution (left) with parameters: padding 1, stride 2, and kernel size 3, compared to our modified convolution (right). Note that the proposed change in input processing does not change the kernel weights.}
    \label{fig:PrAViC_conv3d}
    \vspace{15pt}
\end{figure}

\paragraph{Architecture for Fine-tuning} We outline the modifications made to the classic 3D CNN architecture to adapt it to our approach. While most components of a standard CNN architecture remain unchanged, we specifically alter the 3D convolution processing, batch normalization, and layer pooling. For 3D convolutions, we modify only those layers where the kernel size responsible for the depth (i.e., processing movie frames) is greater than 1. Our modification ensures that the kernels do not extend to the last deep channel. To achieve this, we replicate the input boundary on the front side before performing the multiplication operation, as illustrated in Fig.~\ref{fig:PrAViC_conv3d}. For pooling layers, our modification involves replicating only the first depth channel.

The next essential mechanism in CNN networks is batch normalization~\citep{ioffe2015batch}. It involves calculating the mean and standard deviation for individual dimensions within mini-batches and training gamma and beta parameters during network training. This process ensures consistent and stable training across the network by maintaining the integrity of feature scaling and normalization, which is crucial for effective spatiotemporal pattern learning. In our approach, we modify this mechanism by statically determining the number of depth channels from which the statistics (mean and standard deviation) will be calculated. These statistics are then used to transform the remaining depth channels.

It is important to note that through these operations, each layer of our model only retains information from the preceding depth channels.  In addition, our approach has a unique property: when provided with inputs of the same video, one containing $k$ frames and the other $n$ frames ($k<n$, where the second input is an extension of the first by $n-k$ frames), the network produces identical outputs, constrained to the dimensions of the output of the first $k$ depth channels of the inputs.

\begin{figure}[thb]
    \centering
    \includegraphics[width=\linewidth]{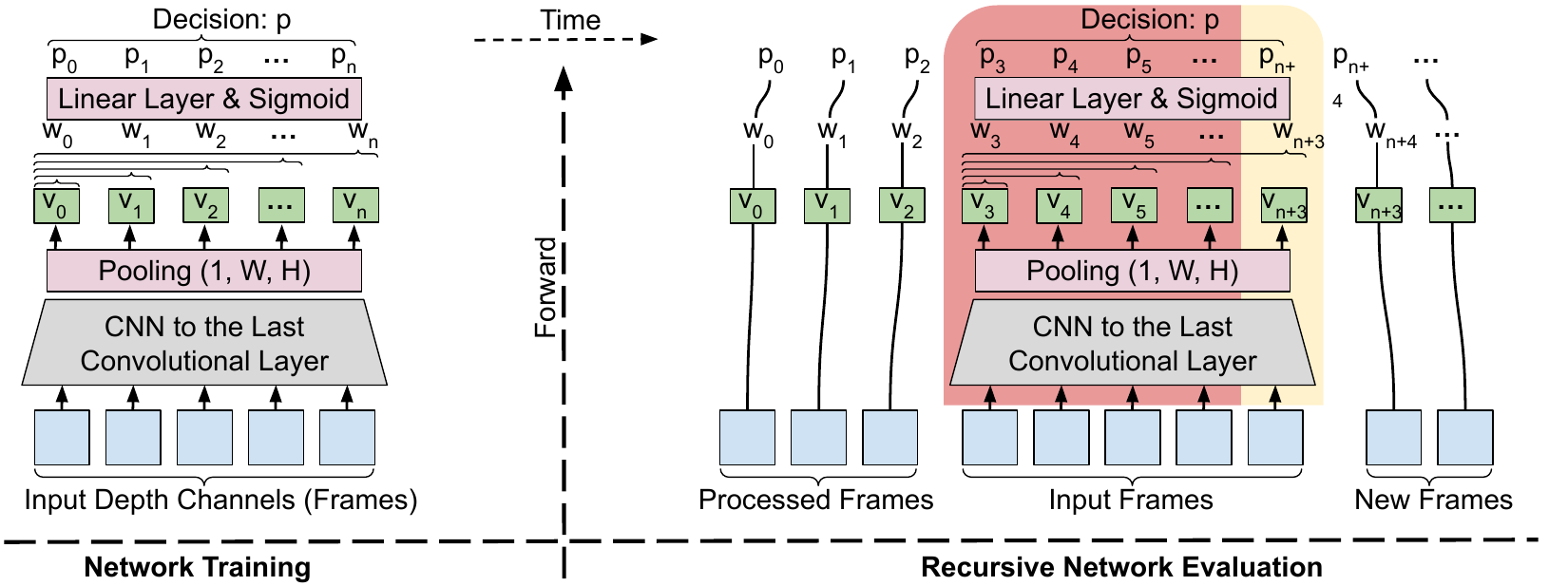}
    \caption{Illustration of our approach during the training phase (left) and the recursive evaluation phase (right). During evaluation, the red area retains computations from previous frames in individual network layers. In contrast, the yellow area represents computations for the recently introduced frame. Without retaining the computations in the red area, they would have to be recalculated, which would lengthen the evaluation process. This approach allows fast online processing of new frames.}
    \label{fig:PrAViC_recursive_evaluation}
    \vspace{15pt}
\end{figure}

\paragraph{Head Design} Here we describe the modifications to the head of the CNN network that allow for online training and recursive evaluation, as shown in Fig.~\ref{fig:PrAViC_recursive_evaluation}. Typically, in the case of a CNN, the head consists of the last linear layer, so in our approach we leave the head as a linear layer, but we will process the output of the last convolutional layer differently than in a classical CNN network. We assume that $v_0,\ldots, v_n\in\mathbb{R}^D$ represent the outputs of the last convolutional and pooling layers, considering only the height and width dimensions, while keeping the depth dimension as it is after the last convolutional layer. With this representation, following the standard offline approach, we perform a mean aggregation of the representations relative to time $t\in\{0,\ldots, n\}$: 
\begin{equation}\label{eq:wt}
\begin{array}{c}
w_t=\tfrac{1}{t+1}\sum\limits_{i=0}^t v_i.
\end{array}
\end{equation}
Using the aggregations $w_t$ above, we process each one separately through a linear layer $h$ followed by the sigmoid function $\sigma$ to obtain $p_t=\sigma(h(w_t))$. The final decision of the model is determined by the formula $p=\max\limits_{t=0..n}p_t$.

%%%%%%%%%%%%%%%%%%%%%%%%%%%%%%

\section{Experiments}\label{sec:experiments}

In this section, the experimental setup is detailed and the results that validate the effectiveness of the proposed approach are presented. Five datasets were employed for testing purposes, comprising four publicly available datasets (UCF101~\citep{soomro2012ucf101},  EgoGesture~\citep{zhang2018egogesture}, Jester~\citep{materzynska2019jester}, and Kinetics-400~\citep{kay2017kinetics}), as well as one closed-access dataset (ultrasound dataset). The experiments employed a variety of architectural approaches, which are described in detail in the corresponding paragraphs, and were conducted using two different types of GPUs: NVIDIA RTX 4090 and NVIDIA A100 40GB. The source code is available at GitHub repository: \url{https://github.com/mtredowicz/PrAViC}.

\paragraph{\our{} vs. Offline Baselines}
\label{sec:comparison_UCF101}
In this paragraph, we undertake a comparative analysis of \our{} with two non-online baseline approaches, namely ResNet-3D-18 (R3D-18)~\citep{tran2018closer} and Separable-3D-CNN (S3D)~\citep{xie2018rethinking}. Experiments were conducted on the widely used video benchmark dataset UCF101~\citep{soomro2012ucf101}, consisting of 13,320 video clips, which are classified into 101 categories.

From each video clip, 16 consecutive frames were extracted, starting with a random one. The R3D-18 model was resized to a resolution of $128\!\times\! 171$, with each frame randomly cropped to $112\!\times\! 112$. In contrast, the S3D model was resized to $128\!\times\! 256$, with each frame randomly cropped to a size of $224\!\times\! 224$.

Initially, the R3D-18 and S3D models underwent pre-training on Kinetics-400. Subsequently, they were modified in accordance with the specifications outlined in Section~\ref{section:architecture}, with the objective of transforming the offline models into online ones. In both cases, we employed Stochastic Gradient Descent (SGD) as the optimizer and Cross-Entropy as the loss function. The learning rates were set to 0.0002 for both of the offline models and to 0.002 and 0.0001 for the modified R3D-18 and S3D models, respectively. Lastly, a custom loss function, as defined in Eq.~\eqref{eq:loss_function_mulit}, was applied with the intention of forcing the model to make an earlier decision. The effectiveness of varying values for the $\lambda$ parameter, starting with 0.1 and increasing to 1, was evaluated. Note that in this case the model is trained to recognize one of the $n=101$ classes.

\begin{figure}
    % \vspace{-15pt}
    \centering
    \includegraphics[width=\linewidth]{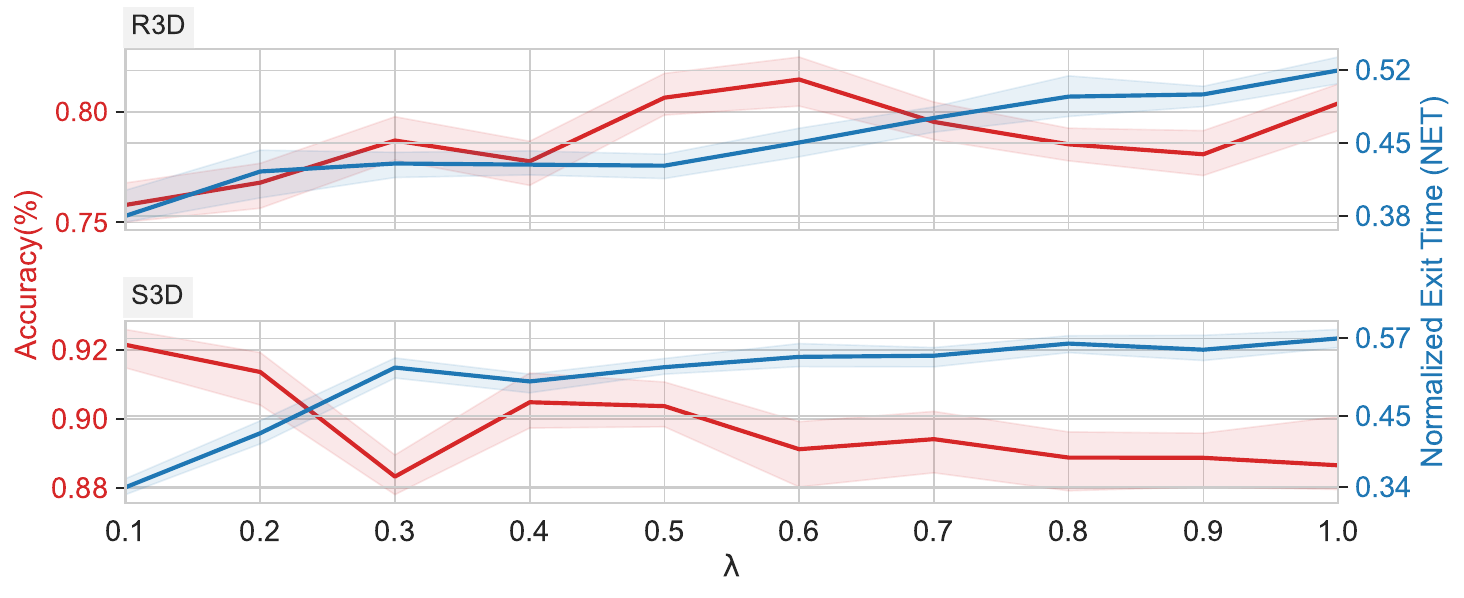}
    \caption{The figure compares accuracy and normalized exit time (NET) as functions of the custom loss function parameter $\lambda$. Results are presented for two base models: R3D-18 (top row) and S3D (bottom row), for which architectures were modified as described in Section~\ref{section:architecture}. Solid lines represent the average values over 30 evaluation trials for each setting, while shaded regions illustrate the full range of observed values (from min to max).
    Note that \our{} tends to delay decisions as $\lambda$ parameter values increase, in contrast to faster decisions observed for lower parameter values. For the R3D-18 baseline, both accuracy and normalized exit time (NET) generally increase with $\lambda$, though accuracy exhibits some fluctuations. In contrast, for the S3D baseline, accuracy generally decreases as $\lambda$ increases, while NET increases consistently. The base models, R3D-18 and S3D, without any changes in architectures and standard Cross-Entropy loss, achieved accuracies of $94.41\%$ and $96.45\%$, respectively. Results are reported on the UCF101 dataset.}
    \label{fig:ucf101_lambda}
    \vspace{15pt}
\end{figure}

Figure~\ref{fig:ucf101_lambda} presents the accuracy and normalized exit time (NET) obtained by the \our{}$_\lambda$ modification (for the $\lambda$ parameter varying from 0.1 up to 1) applied to the considered baselines, namely R3D-18 and S3D, which architectures were modified as described in Section~\ref{section:architecture}.

It has been observed that as the parameter $\lambda$ increases, the normalized exit time exhibits a clear increasing trend across both models, indicating that higher values of $\lambda$ encourage later exits. In contrast, accuracy does not follow a consistent trend -- while it generally increases for R3D-18, it fluctuates, and for S3D, it tends to decrease with higher $\lambda$. This suggests that the impact of $\lambda$ on accuracy is less predictable than its effect on NET. 

Nonetheless, the results that have been presented do not delve into the relationship between accuracy (or error rate) and normalized exit time. Therefore, a respective ablation study is required (see below). The corresponding significance analysis can be found in the Supplementary Material~\cite{trędowicz2024pravicprobabilisticadaptationframework}.

\paragraph{Ablation Study}
To further analyze the relationship between error rate and normalized exit time (NET), we conduct an ablation study on the performance of the two base models, namely R3D-18 and S3D, using various \our{}$_{\lambda}$ configurations, where $\lambda$ varies from 0.1 to 1. This analysis aims to highlight the trade-offs between error rates and early exits efficiency by evaluating how different $\lambda$ values influence model's behavior. 

\begin{figure}[thb]
    \centering
    \includegraphics[width=\linewidth]{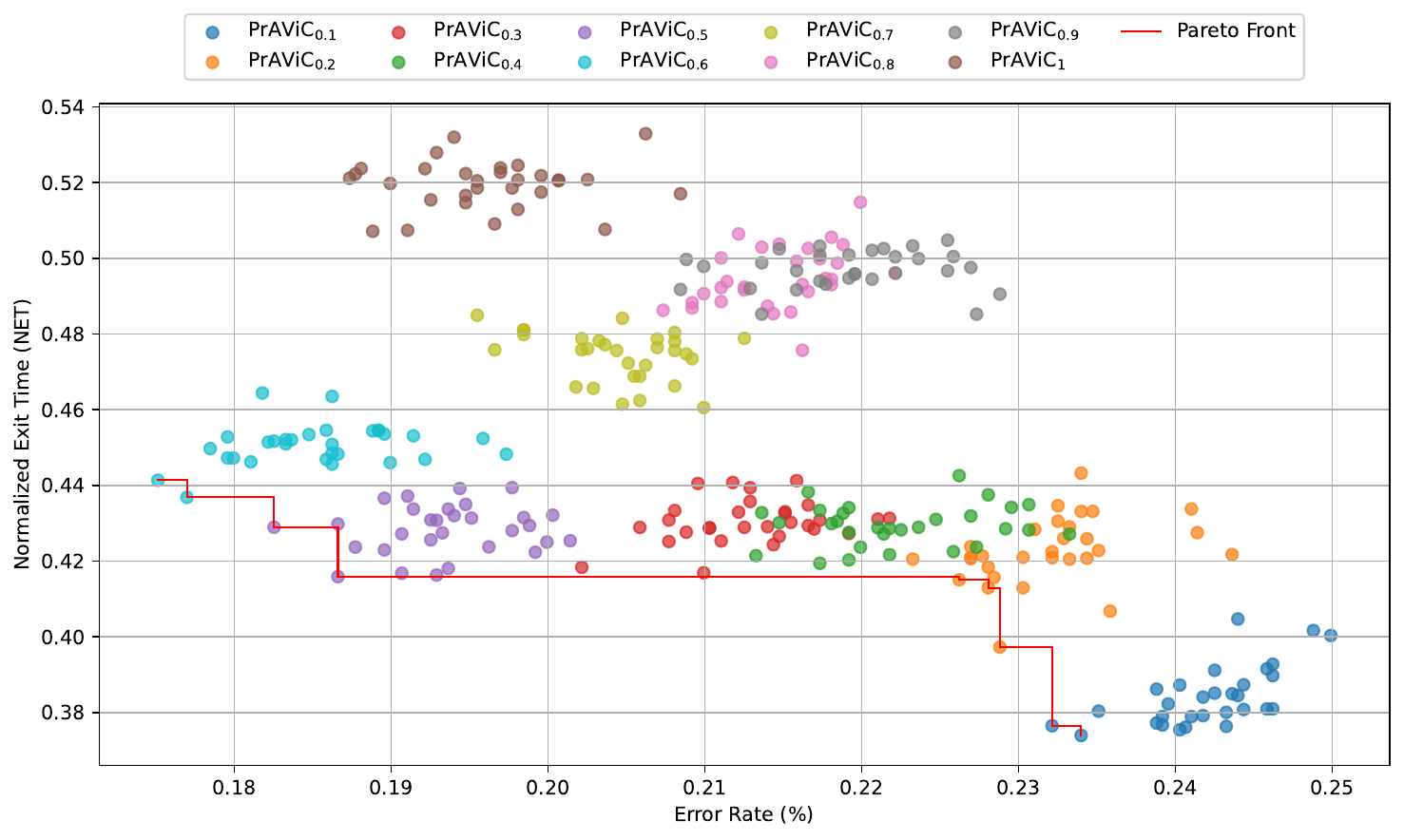}
    \caption{Performance comparison of different \our{}$_\lambda$ methods, where $\lambda$ ranges from 0.1 to 1, with the R3D-18 baseline model. The Pareto front highlights the optimal trade-offs between normalized exit time (NET) and error rate achieved across the evaluated configurations. Note that Pareto-optimal solutions consist exclusively of points corresponding to  $\lambda$ values ranging from 0.1 to 0.6.}
    \label{fig:PrAViC_pareto_front_r3d}
    \vspace{15pt}
\end{figure}

\begin{figure}[thb]
    \centering
    \includegraphics[width=\linewidth]{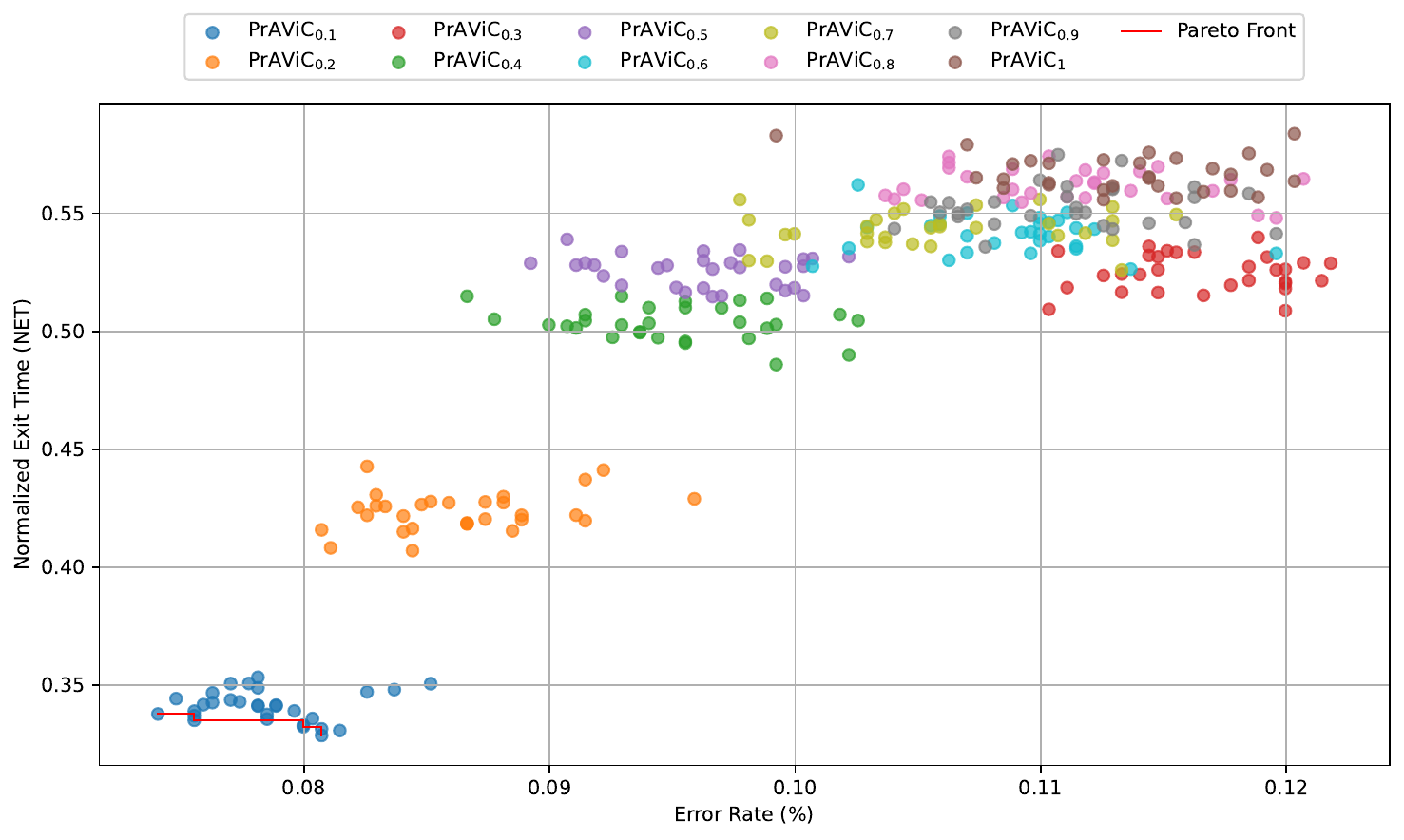}
    \caption{Performance comparison of different \our{}$_\lambda$ methods, where $\lambda$ ranges from 0.1 to 1, with the S3D baseline model. The Pareto front highlights the optimal trade-offs between normalized exit time (NET) and error rate achieved across the evaluated configurations. Note that Pareto-optimal solutions consist exclusively of points corresponding to the setting $\lambda = 0.1$.}
    \label{fig:PrAViC_pareto_front_s3d}
    \vspace{15pt}
\end{figure}

Scatter plots presented in Fig.~\ref{fig:PrAViC_pareto_front_r3d} and Fig.~\ref{fig:PrAViC_pareto_front_s3d} illustrate the distribution of the error rate versus NET for different \our{} configurations, for R3D-18 (Fig.~\ref{fig:PrAViC_pareto_front_r3d}) and S3D (Fig.~\ref{fig:PrAViC_pareto_front_s3d}) baseline models, respectively. Each colored cluster of points corresponds to a specific $\lambda$ value, as indicated in the legend. For each $\lambda$ value, the model's performance in terms of error rate and NET is evaluated 30 times using different random seeds, resulting in 30 data points per setting. The x-axis represents the model's error rate in percentage, with lower values indicating better performance (and therefore higher accuracy). The y-axis shows the normalized exit time (NET), where lower values suggest earlier exit decisions during inference. This visualization provides a detailed perspective on the variability in performance. Observing the distribution of points for the R3D-18 baseline, it can be seen that models with lower $\lambda$ values, such as \our{}$_{0.1}$ and \our{}$_{0.2}$, tend to achieve earlier exit times, while higher $\lambda$ values generally lead to later exit times and lower error rates. On the other hand, for the S3D baseline lower $\lambda$ values, such as \our{}$_{0.1}$ and \our{}$_{0.2}$, tend to result in lower error rates and earlier exit times, whereas higher $\lambda$ values generally lead to extended exit times and potentially increase error rate.

We also include Pareto fronts that identify the optimal balance between NET and error rate across the tested settings. Points connected by the solid red line on the plot are Pareto-optimal solutions, with no other solutions better in both objectives -- error rate and normalized exit time. Notably, for both of the baseline models, the Pareto Front consists exclusively of points corresponding to lower $\lambda$ values (ranging from 0.1 to 0.6 for the R3D-18 model and 0.1 for the S3D model), suggesting that these configurations achieve the most favorable trade-offs between error rate and normalized exit time. This justifies that our method of encouraging the model to make early decisions provides, to the same extent, optimal results.

\paragraph{Significance Analysis for Experiments on the UCF101 Dataset}\label{app:confidence}

We performed a significance analysis following our experimental study involving \our{} with the R3D-18 and S3D baselines, conducted on the UCF101 dataset. To this end, we employed the models with varying values of the parameter $\lambda$ as classifiers and subsequently compared them through the utilization of the methodology for the comparison of multiple models, which is based on the Friedman pair-wise ranking tests with the Conover post-hoc test \cite{conover1979multiple}. In each case, we established a significance level of 0.05.

\begin{figure*}[thb]
    \centering
    \includegraphics[width=0.9\linewidth]{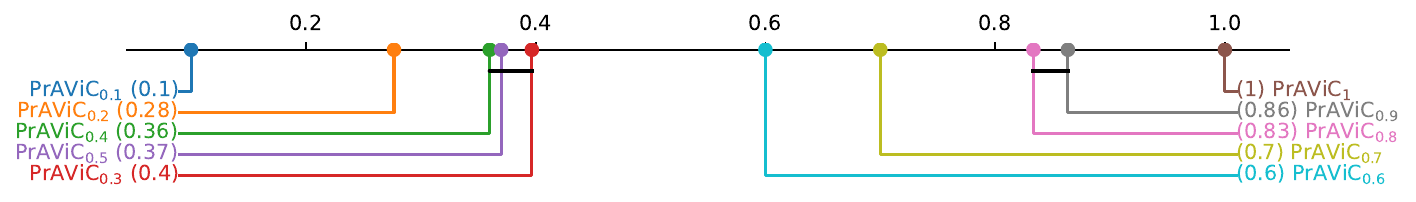}
    \caption{Statistical comparison of several \our{} models with the R3D-18 baseline, trained on the UFC101 dataset. The horizontal axis represents the normalized exit time (NET) and each colored line represents a different \our{}$_\lambda$ method, where $\lambda$ ranges from 0.1 to 1. The corresponding average normalized exit times are given in parentheses and plotted along the horizontal axis. When methods are connected by a black line, they perform similarly within statistical confidence limits. Methods without connections to others can be considered significantly different from the rest. Note that the best performing methods (lowest NET) are \our{}$_{0.1}$, \our{}$_{0.2}$, and \our{}$_{0.4}$. On the other hand, the worst performing methods are \our{}$_{0.8}$, \our{}$_{0.9}$, and \our{}$_{1}$.}
    \label{fig:PrAViC_statistical_critical_diagram_net_r3d}
    \vspace{15pt}
\end{figure*}

\begin{figure*}[thb]
    \centering
    \includegraphics[width=0.9\linewidth]{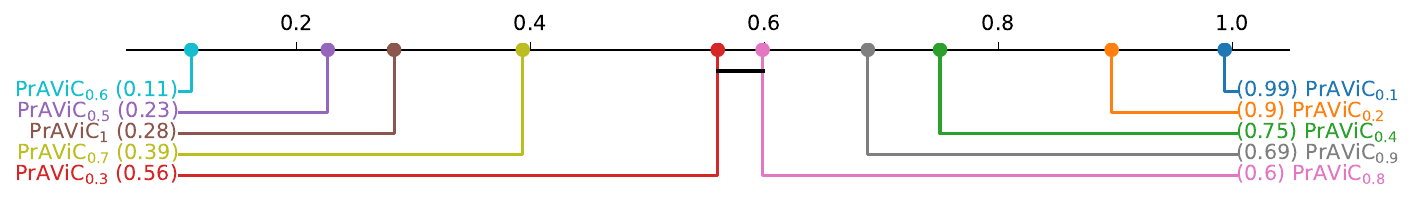}
    \caption{Statistical comparison of several \our{} models with the R3D-18 baseline, trained on the UFC101 dataset. The horizontal axis represents the test error rate and each colored line represents a different \our{}$_\lambda$ method, where $\lambda$ ranges from 0.1 to 1. The corresponding average test error values are given in parentheses and plotted along the horizontal axis. When methods are connected by a black line, they perform similarly within statistical confidence limits. Methods without connections to others can be considered significantly different from the rest. Note that the best performing methods (lowest test error rate) are \our{}$_{0.6}$, \our{}$_{0.5}$, and \our{}$_{1}$. On the other hand, the worst performing methods are \our{}$_{0.1}$ and \our{}$_{0.2}$.}
    \label{fig:PrAViC_statistical_critical_diagram_test_error_r3d}
    \vspace{15pt}
\end{figure*}

The obtained results, together with their detailed descriptions, are presented in Figures \ref{fig:PrAViC_statistical_critical_diagram_net_r3d}, \ref{fig:PrAViC_statistical_critical_diagram_test_error_r3d}, for the \our{} models with the R3D-18 baseline and in the Supplementary Material~\cite{trędowicz2024pravicprobabilisticadaptationframework} for the \our{} models with the S3D baseline. However, it should be emphasized that we can also make two general observations.  First, note that most of the \our{} models considered give statistically different results. Second, the relationship between normalized exit times and error rates depends on the base model used. While for the R3D-18 model certain values of $\lambda$ are unlikely to favor both NET and accuracy, this often happens for the S3D model. Note that this is consistent with our previous results presented in Section~\ref{sec:experiments} (see Figures \ref{fig:ucf101_lambda}, \ref{fig:PrAViC_pareto_front_r3d}, and \ref{fig:PrAViC_pareto_front_s3d}).

\paragraph{\our{} vs. {\em Co} Models} 
We conducted experiments using the {\em Co}X3D models introduced in \cite{hedegaard2022continual}, preserving the identical experimental setting. Our modification of the {\em Co}X3D network (variants S, M, and L) was evaluated on the test set of the Kinetics-400 dataset~\citep{kay2017kinetics}, which contains 400 human action classes, with at least 400 video clips for each action. One temporally centered clip was extracted from each video. The publicly accessible weights were utilized without any additional fine-tuning. The loss function from Eq.~\eqref{eq:loss_function_mulit} was implemented for $n=400$ classes.

\begin{table}[tbh]\small
\centering
\vspace{10pt}
\caption{The performance of the {\em Co}X3D and \our{} X3D-based models trained on the Kinetics-400 dataset is evaluated. The accuracy and normalized exit time (NET) are calculated for the test set. It is observed that our solution, which aggregates the temporal information in a manner relative to time, exhibits superior performance.}
\vspace{15pt}
\resizebox{.8\linewidth}{!}{%
\begin{tabular}{@{}c@{\quad}c@{\;\,}c@{\;\,}c@{\;\,}c@{\;\,}c@{}}
\toprule
\multirow{2}{*}{\textbf{X3D Variant}} & \multicolumn{2}{c}{\textbf{{\em Co} Model}} & \multicolumn{2}{c}{\textbf{\our{$_1$} Model}} \\
\cmidrule(lr){2-3} \cmidrule(lr){4-5}
 & Accuracy (\%) & NET & Accuracy (\%) & NET \\
\midrule  
X3D-S\textsubscript{64} & 67.33 & 1 & \textbf{67.60} & 0.03 \\
% \midrule 
X3D-S\textsubscript{13}  & 60.18 & 1 & \textbf{66.43} & 0.27\\
% \midrule 
X3D-M\textsubscript{64}  & 71.03 & 1 & \textbf{72.92} & 0.17\\
% \midrule 
X3D-M\textsubscript{16}  & 62.80 & 1 & \textbf{70.42} & 0.59 \\
% \midrule 
X3D-L\textsubscript{64} & 71.61 & 1 & \textbf{73.47} & 0.03 \\
% \midrule 
X3D-L\textsubscript{16}  & 63.03 & 1 & \textbf{69.05} & 0.27 \\
\bottomrule
\end{tabular}
}
\label{tab:kinetic400}
\end{table}

The comparison results, comprising accuracy scores and normalized exit times (NETs), for the {\em Co}X3D and \our{}$_1$ X3D-based models are presented in Tab.~\ref{tab:kinetic400}. It should be noted that while the {\em Co}X3D models always make a decision on the last frame, \our{}$_1$ (though not particularly forced for early exit) allows for the possibility of making a decision earlier with higher accuracy. 
% The objective of this paragraph is to further investigate the influence of our method on the accuracy of the model and the exit time. 
This justifies the ability of our method, by simply aggregating information across frames in a more structured way (via specific head design), to improve the model's capability to make timely decisions while maintaining or improving accuracy.

\paragraph{\our{} vs. Online Models}

This paragraph presents a comparison between \our{} and a collection of state-of-the-art online models, including 3D-SqueezeNet, 3D-ShuffleNet, 3D-ShuffleNetV2, and 3D-MobileNetV2 \citep{kopuklu2020online}. The models were trained on the EgoGesture~\citep{zhang2018egogesture} and Jester~\citep{materzynska2019jester} datasets, which have been specifically designed for the purpose of recognizing hand gestures from an egocentric perspective, and cover 83 (EgoGesture) and 27 (Jester) different classes, respectively.
%This dataset serves as a robust evaluation platform, catering to both segmented gesture classification and continuous gesture detection. It features 83 classes of static and dynamic gestures captured across 6 diverse indoor and outdoor environments. The dataset partitions, created by different contributors, adhere to a 3:1:1 ratio, yielding 1239 training videos, 411 validation videos, and 431 testing videos, containing 14416, 4768, and 4977 gesture samples, respectively. Data is available in two formats: RGB (full color) and depth module (grayscale, emphasizing hand gestures). Our experiments focused exclusively on the depth module data type.

In setting up these experiments, the learned model weights (see \citep{kopuklu2020online}) were utilized on the data. Initially, the model heads (of which there are 83 or 27, respectively, depending on the dataset) were trained anew, followed by light retraining of the entire model, with or without incorporation of an additional cost function provided in Eq.~\eqref{eq:loss_function_mulit} to promote earlier detection. The architectures employed in our approach were identical to those utilized in the baseline models, with the requisite modifications detailed in Section~\ref{section:architecture}. The models were trained using the Stochastic Gradient Descent (SGD) optimizer with standard Categorical Cross-Entropy loss. The momentum, damping, and weight decay were set to 0.9, 0.9, and 0.001, respectively. The network learning rate was initialized at 0.1, 0.05, and 0.01, then decreased by a factor of 3 with a factor of 0.1 when the validation loss reached convergence.

During the training phase, input clips were selected from random time points within the video clip. If the video was composed of a smaller number of frames than the specified input size, a loop padding was incorporated. For the purpose of inputting data into the network, clips comprising multiples of 16 frames were utilized. In particular, 128 frames were used for 3D-MobileNetV2, while the remaining models utilized 192 frames. This approach was adopted to yield 8 probability values $p_i$, at the network output for the MobileNet network and 12 for the other networks. A single input clip possessed dimensions of $3\!\times\! k\!\times\! 112\!\times\! 112$, where $k$ represents the number of frames, which varied based on the models used (as described above).

\begin{table}[thb]
\centering
\caption{Comparison of top-1 (ACC@1) and top-5 (ACC@5) accuracy scores for \our{} and baseline models (i.e., 3D-SqueezeNet, 3D-ShuffleNet, 3D-ShuffleNetV2, and 3D-MobileNetV2) trained on the EgoGesture and Jester datasets, demonstrating the improved performance of our approach and supporting online recursive exploitation.}
\vspace{15pt}
\label{tab:performance}
\resizebox{\linewidth}{!}{%
\begin{tabular}{@{}c@{}c@{\quad}l@{}r@{\quad}l@{}r@{\quad}l@{}r@{\quad}r@{\!\!}r@{}}
% \toprule
\cmidrule[.75pt](l{-10pt}r{-1pt}){2-10}
& \textbf{Baseline:} & \multicolumn{2}{l}{\textbf{SqueezeNet}} & \multicolumn{2}{l}{\textbf{ShuffleNet}} & \multicolumn{2}{l}{\textbf{ShuffleNetV2}} & \multicolumn{2}{l}{\textbf{MobileNetV2}} \\ 
\cmidrule(r){3-4} \cmidrule(r){5-6} \cmidrule(r){7-8} \cmidrule(r){9-10}
 &  & Base & \our{$_1$} & Base & \our{$_1$} & Base & \our{$_1$} & Base & \our{$_1$} \\[2pt] 
% \midrule
\cmidrule[.55pt](l{-10pt}r{-1pt}){2-10}
\multirow{2}{*}{\rotatebox{45}{\textit{EgoGesture}}} & ACC@1  & 88.23 & \textbf{91.84} & 89.93 & \textbf{99.04} & 90.44 & \textbf{98.49} & 90.31 & \textbf{99.24} \\ 
\cmidrule(lr){2-10}
& ACC@5 & 97.63 & \textbf{99.20} & 98.28 & \textbf{99.73} & 98.36 & \textbf{99.16} & 98.22 & \textbf{99.79} \\
\cmidrule(l{-10pt}r{-1pt}){2-10}
% \midrule
\multirow{3}{*}{\rotatebox{45}{\textit{Jester}}} & ACC@1 & \textbf{90.74} & 90.52 & 93.08 & \textbf{94.48} & 93.69 & \textbf{95.45} & 94.34 & \textbf{95.25} \\ 
\cmidrule(lr){2-10}
& ACC@5 & \textbf{96.75} & 95.99 & 99.50 & \textbf{99.53} & 99.57 & \textbf{99.59} & \textbf{99.63} & 99.62 \\[2pt] 
\cmidrule[.75pt](l{-10pt}r{-1pt}){2-10}
% \bottomrule
\end{tabular}%
}
\end{table}

\begin{table}[thb]
\centering
\caption{Top-1 (ACC@1) and top-5 (ACC@5) accuracy scores and normalized exit times (NETs) for \our{} with and without the early decision cost function term (i.e., with the parameter $\lambda$ equal to 1 and 0.9), trained for various baseline models (i.e., 3D-SqueezeNet, 3D-ShuffleNet, 3D-ShuffleNetV2, and 3D-MobileNetV2) on the EgoGesture and Jester datasets. Note that penalizing late decisions leads to reduced exit time without significantly affecting accuracy.}
\vspace{15pt}
\label{tab:efficiency_of_our}
\resizebox{1\linewidth}{!}{%
\begin{tabular}{@{}c@{\!}c@{\quad}r@{}r@{\;\,}r@{}r@{\;\,}r@{}r@{\;\,}r@{}r@{}}
% \toprule
\cmidrule[.75pt](l{-10pt}r{-1pt}){2-10}
& \textbf{Baseline:} & \multicolumn{2}{c}{\textbf{SqueezeNet}} & \multicolumn{2}{c}{\textbf{ShuffleNet}} & \multicolumn{2}{c}{\textbf{ShuffleNetV2}} & \multicolumn{2}{c}{\textbf{MobileNetV2}} \\ 
\cmidrule(r){3-4} \cmidrule(r){5-6} \cmidrule(r){7-8} \cmidrule(r){9-10}
 &  & \our{$_1$} & \our{$_{0.9}$} & \our{$_1$} & \our{$_{0.9}$} & \our{$_1$} & \our{$_{0.9}$} & \our{$_1$} & \our{$_{0.9}$} \\[2pt] 
% \midrule
\cmidrule[.55pt](l{-10pt}r{-1pt}){2-10}
\multirow{3}{*}{\rotatebox{70}{\textit{EgoGesture}}} & NET & 0.5 & 0.3 & 0.7 & 0.4 & 0.7 & 0.4 & 0.6 & 0.1 \\
\cmidrule(lr){2-10}
 & ACC@1 & \textbf{91.84} & 90.86 & \textbf{99.04} & 98.70 & \textbf{98.49} & 97.00 & 99.24 & \textbf{99.58} \\ 
\cmidrule(lr){2-10}
& ACC@5 & \textbf{99.20} & 99.16 & \textbf{99.73} & 99.66 & \textbf{99.16} & 99.08 & 99.79 & \textbf{99.92} \\
\cmidrule(l{-10pt}r{-1pt}){2-10}
% \midrule
\multirow{4}{*}{\rotatebox{70}{\textit{Jester}}} & NET & 0.2 & 0.05 & 0.6 & 0.25 & 0.55 & 0.05 & 0.5 & 0.05 \\
\cmidrule(lr){2-10}
& ACC@1 & \textbf{90.52} & 90.51 & \textbf{94.48} & 94.18 & \textbf{95.45} & 95.03 & 95.25 & \textbf{95.35} \\ 
\cmidrule(lr){2-10}
& ACC@5 & \textbf{95.99} & 95.94 & \textbf{99.53} & 99.03 & \textbf{99.59} & 99.55 & 99.62 & \textbf{99.66} \\[2pt] 
\cmidrule[.75pt](l{-10pt}r{-1pt}){2-10}
% \bottomrule
\end{tabular}%
}
\end{table}

Tab.~\ref{tab:performance} provides a detailed comparison of the performance of \our{} and the baseline models  (i.e., 3D-SqueezeNet, 3D-ShuffleNet, 3D-ShuffleNetV2, and 3D-MobileNetV2) trained on the EgoGesture and Jester datasets, with the presentation of the top-1 and top-5 accuracy results for the test set. As can be seen, \our{} generally outperforms the other models across most metrics. 
% Furthermore, our approach offers the additional benefit of being applicable in an online, recursive manner, providing enhanced functionality and efficiency in practical applications (see Fig.~\ref{fig:PrAViC_recursive_evaluation} for a visual representation). 
These results demonstrate the efficacy of our approach and highlight its potential to deliver superior performance in online gesture recognition tasks.
Additionally, Tab.~\ref{tab:efficiency_of_our} shows the results of our model with and without a cost function term for early decisions (see Eq.~\eqref{eq:loss_function_mulit}). We can observe that penalizing late decisions leads to reduced exit time without significantly affecting accuracy.
% \paragraph{Ablation studies}

\paragraph{\our{} for Medical Use}

In our final experiment, for which details are presented in the Supplementary Material~\cite{trędowicz2024pravicprobabilisticadaptationframework}, we tested \our{} on real-life dataset of Doppler ultrasound videos representing short-axis and suprasternal views of newborn hearts. The task was binary classification, where the goal was to predict whether congenital heart disease occurs or not.  Pre-trained offline 3D-CNN baseline model was modified, as described in Section~\ref{section:architecture}. Although the model's accuracy decreased slightly (from $94\%$ for the offline model to approximately $90\%$ for \our{}), we successfully developed a model that reaches a final decision expeditiously (see the Supplementary Material~\cite{trędowicz2024pravicprobabilisticadaptationframework}).
% Despite a slight reduction in accuracy (approximately $90\%$) in comparison to the offline model ($94\%$), we were able to develop a model that reaches a final decision expeditiously (see Tab.~\ref{tab:ultrasound} in Appendix~\ref{app:ultra}).

%%%%%%%%%%%%%%%%%%%%%%%%%%%%%%

\section{Conclusions}

In this work, we proposed \our{}, a general framework for automatically modifying networks that have been adapted for video processing to their online counterparts. The objective of \our{} was to identify potentially dangerous situations at the earliest possible stage. To achieve this, we introduced a probabilistic theoretical model that underlies online data processing. We computed the mean expected exit time and used it as a component of the loss function to encourage the model to make early decisions. Furthermore, we proposed a simple framework that translates offline models into online counterparts, based on a specific customization of the head design. In a series of experiments conducted, it was shown that \our{} can encourage the model to make decisions earlier without a significant decrease in accuracy.

\paragraph{Limitations}
The adaptations implemented by \our{} do not extend to architectures such as transformers or CNN-LSTMs. In fact, standard convolutions with a time component are employed.  While this enables the utilization of pre-trained networks, it does not permit the comprehensive utilization of the full frames of the video in the residual network. This may potentially result in a minor loss of accuracy in comparison to the offline model.

\section*{Impact Statement}

As the use of online video analysis becomes increasingly prevalent in both medical and social contexts, our methodology can be employed as a straightforward instrument for researchers engaged in applied video processing. Moreover, as this paper presents work whose goal is to advance the field of machine learning, there are many other potential societal consequences of our work, none of which we feel must be specifically highlighted here.

\section*{Acknowledgments} This research was partially funded by the National Science Centre, Poland, grants no. 2023/49/B/ST6/01137 (work by Jacek Tabor and \L{}ukasz Struski) and 2023/50/E/ST6/00068 (work by Marcin Mazur). Some experiments were performed on servers purchased with funds from the flagship project entitled ``Artificial Intelligence Computing Center Core Facility'' from the DigiWorld Priority Research Area within the Excellence Initiative -- Research University program at Jagiellonian University in Kraków. The authors would also like to give their acknowledgments to the Prometheus MedTech.AI for financial support of the ongoing research work reported in this paper.

% \section{Citations and references}
\bibliography{ecai2025_conference}

%%%%%%%%%%%%%%%%%%%%%%%%%%%%%%%%%%%%%%%%%%%%%%%%%%%%%%%%%%%%%%%%
%%%%%%%%%%%%%%%%%%%%%%%%%%%%%%%%%%%%%%%%%%%%%%%%%%%%%%%%
%%%%%%%%%%%%%%%%%%%%%%%%%%%%%%%%%%%%%%%%%%%%%
%%%%%%%%%%%%%%%%%%%%%%%%%%%%%%%%%%
%%%%%%%%%%%%%%%%%%%%%%%%%
%%%%%%%%%%%%%%%%%
%%%%%%%%%%
%%%%%%%%
\newpage
\appendix

\begin{strip}
\centering
\rule{10cm}{0.4pt} \\
\Huge\textbf{Supplementary Material} \\[-15pt]
% \medskip
\rule{10cm}{0.4pt}
% \hrule
\vspace{25pt}
\end{strip}

%%%%%%%%%%
%%%%%%%%%%%%%%%%%
%%%%%%%%%%%%%%%%%%%%%%%%%
%%%%%%%%%%%%%%%%%%%%%%%%%%%%%%%%%%

\section{Additional Experimental Results}

This section presents additional results that were not included in the main paper. These supplementary findings provide further insights and strengthen the conclusions drawn from our primary analyses. The results discussed here include statistical analysis and complementary experiments that provide a broader perspective on the research topic. 

\subsection{Significance Analysis for Experiments on the UCF101 Dataset}\label{app:confidence}

We performed a significance analysis following our experimental study involving \our{} with the S3D baseline, conducted on the UCF101 dataset. 

\begin{figure*}[thb]
    \centering
    \includegraphics[width=\linewidth]{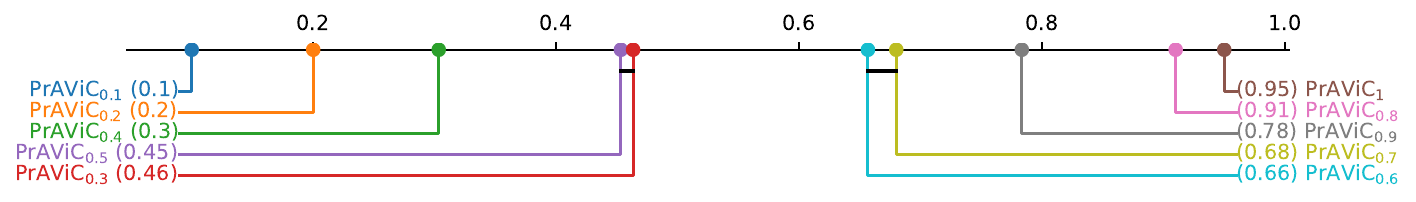}
    \caption{Statistical comparison of several \our{} models with the S3D baseline, trained on the UFC101 dataset. The horizontal axis represents the normalized exit time (NET) and each colored line represents a different \our{}$_\lambda$ method, where $\lambda$ ranges from 0.1 to 1. The corresponding average normalized exit times are given in parentheses and plotted along the horizontal axis. When methods are connected by a black line, they perform similarly within statistical confidence limits. Methods without connections to others can be considered significantly different from the rest. Note that the best performing methods (lowest NET) are \our{}$_{0.1}$, \our{}$_{0.2}$, and \our{}$_{0.4}$. On the other hand, the worst performing methods are \our{}$_{0.8}$, \our{}$_{0.9}$, and \our{}$_{1}$. }
    \label{fig:PrAViC_statistical_critical_diagram_net_s3d}
    \vspace{15pt}
\end{figure*}

\begin{figure*}[thb]
    \centering
    \includegraphics[width=\linewidth]{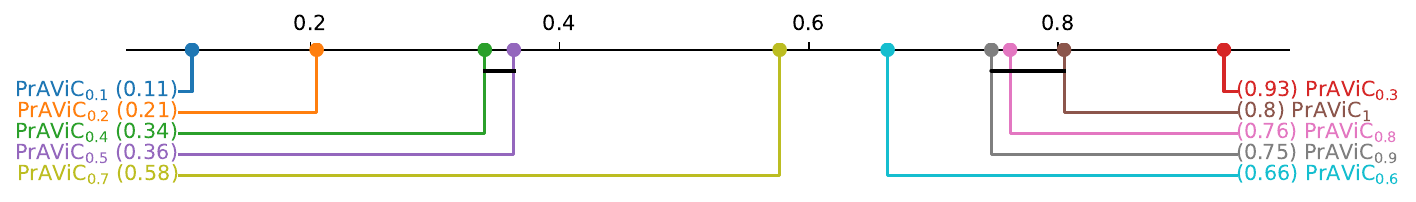}
    \caption{Statistical comparison of several \our{} models with the S3D baseline, trained on the UFC101 dataset. The horizontal axis represents the test error rate and each colored line represents a different \our{}$_\lambda$ method, where $\lambda$ ranges from 0.1 to 1. The corresponding average test error values are given in parentheses and plotted along the horizontal axis. When methods are connected by a black line, they perform similarly within statistical confidence limits. Methods without connections to others can be considered significantly different from the rest. Note that the best performing methods (lowest test error rate) are \our{}$_{0.1}$, \our{}$_{0.2}$, and \our{}$_{0.4}$. On the other hand, the worst performing methods are \our{}$_{0.3}$, \our{}$_{0.8}$, \our{}$_{0.9}$, and \our{}$_{1}$.}
    \label{fig:PrAViC_statistical_critical_diagram_test_error_s3d}
    \vspace{15pt}
\end{figure*}

The obtained results, together with their detailed descriptions, are presented in Figures \ref{fig:PrAViC_statistical_critical_diagram_net_s3d} and \ref{fig:PrAViC_statistical_critical_diagram_test_error_s3d}. Comprehensive results for the R3D baseline, together with their detailed analysis, are provided in the main paper. However, it should be emphasized that we can also make two general observations. First, note that most of the \our{} models considered give statistically different results. Second, the relationship between normalized exit times and error rates depends on the base model used. While for the R3D-18 model certain values of $\lambda$ are unlikely to favor both NET and accuracy, this often happens for the S3D model. Note that this is consistent with our previous results presented in the main paper. 

\subsection{Study of $\NE$ Function During Training}

\begin{figure*}[thb]
    \centering
    \includegraphics[width=0.95\textwidth]{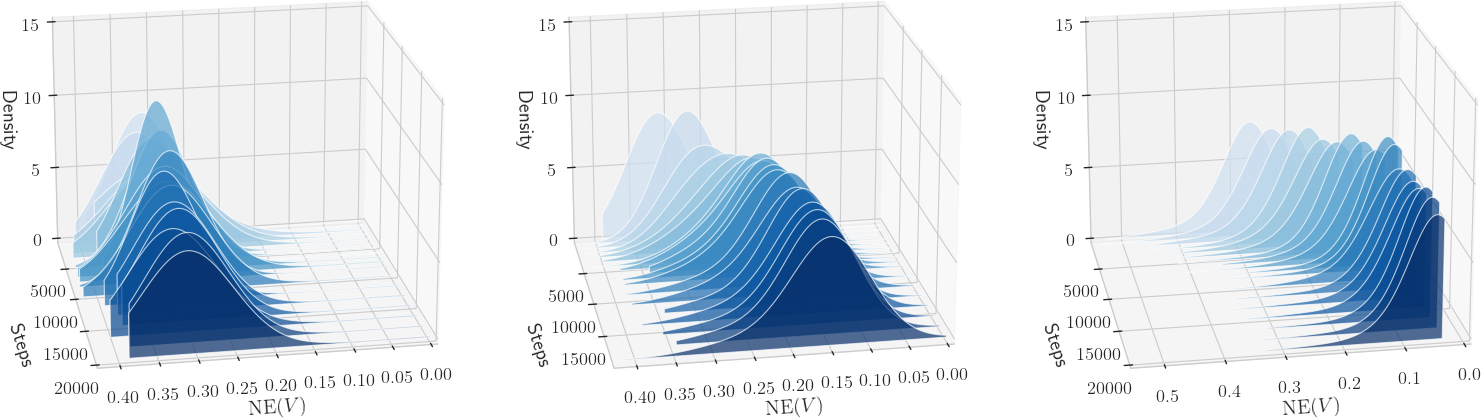}
    \caption{Illustration of the progression of changes in the $\NE$ function during 3D-SqueezeNet training process (from left to right) using the cost function given in the main paper for $\lambda = 1, 0.9, \text{and}\ 0.5$, respectively. For $\lambda = 1$, where late detection of classes is not penalized, the $\NE$ value remains relatively stable. However, as the $\lambda$ parameter decreases, the $\NE$ value shows a more noticeable decline, indicating the increasing impact of penalizing late detection on the model performance.}
    \label{fig:dynamical_add_loss}
    \vspace{15pt}
\end{figure*}

Based on the 3D-SqueezeNet model, we show the course of changes in the $\NE$ function (see the main paper) while training this model. The image demonstrates these changes using the cost function for $\lambda = 1, 0.9, \text{and}\ 0.5$, respectively. As depicted in Fig.~\ref{fig:dynamical_add_loss}, with $\lambda = 1$, the $\NE$ value remains relatively stable, whereas a reduction in the $\lambda$ parameter results in a more noticeable decline in the $\NE$ value, highlighting the effect of penalizing late detection on model performance.

\begin{figure*}[thb]
    \centering
    \includegraphics[width=0.95\textwidth]{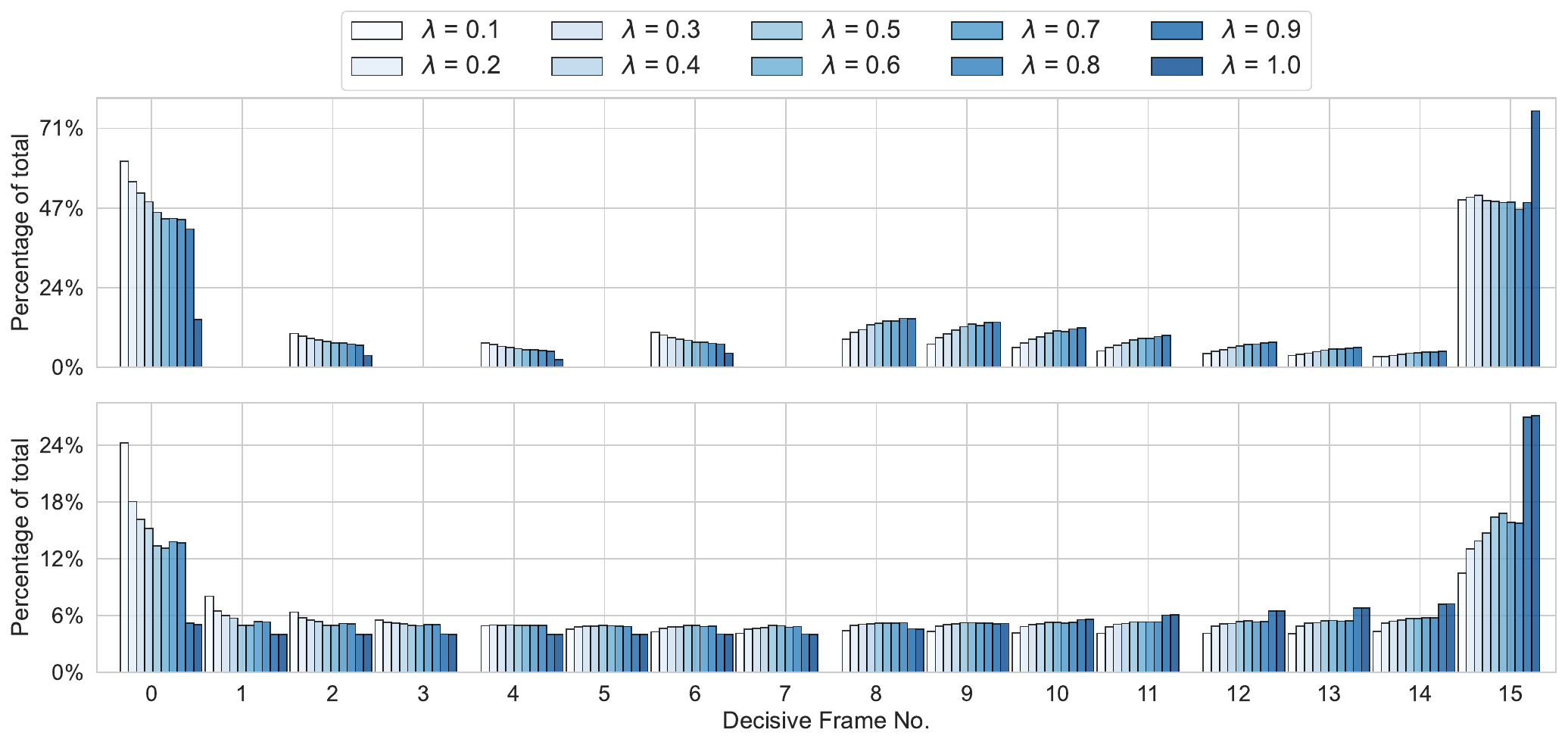}
    \caption{Comparison of histograms of decisive frame numbers for the R3D-18-based (top row) and S3D-based (bottom row) \our{} models with different loss function $\lambda$ parameters, varying from 0.1 to 0.9. The higher the $\lambda$ value, the higher the percentage of higher numbers of decisive frames. Decreasing the $\lambda$ parameter encourages the model to make decisions earlier.}
    \label{fig:ucf101_hist}
    \vspace{15pt}
\end{figure*}

Fig.~\ref{fig:ucf101_hist} presents histograms of decisive frame numbers for \our{$_\lambda$} with the R3D-18 (top row) and S3D (bottom row) baseline models (see the main paper), comprising different values of the $\lambda$ parameter. In both cases, we observe that larger $\lambda$ values (depicted in deep blue shades) result in higher decisive frame numbers. Conversely, smaller $\lambda$ values (depicted in white and light blue shades) lead the model to make decisions much earlier, as expected.

\subsection{Experiments on the Doppler Ultrasound Dataset}\label{app:ultra}

The aim of experiments presented in this section is to demonstrate the efficacy of the \our{} in identifying congenital heart defects (CHDs) in neonates through ultrasound imaging. Congenital heart defects, encompassing conditions such as Tetralogy of Fallot, Hypoplastic Left Heart Syndrome, and Ventricular Septal Defect, pose formidable diagnostic hurdles owing to their intricate nature and the subtleties inherent in early cardiac anomalies. Regrettably, undetected instances of such defects represent a prominent contributor to neonatal mortality rates. In medical practice, however, it is not only crucial to achieve high accuracy but also to make timely decisions, as early intervention can significantly improve the outcomes for affected infants.

\paragraph{\our{} vs. Base Model} We conducted an experiment, in which \our{} was tested on a real medical dataset of Doppler ultrasound videos representing short-axis and suprasternal views of newborns' hearts. These recordings were obtained as part of an ongoing scientific research project involving pediatric cardiologists, with the consent of the newborns' parents. During the acquisition process, a total of 18,365 ultrasound recordings were collected. 
%However, only Doppler effect projections of varying lengths and resolutions were selected for the experiment. In the preprocessing stage, they were transformed to a resolution of $200\!\times\!200$ and then randomly croped to size $172\!\times\!172$. The contrast and brightness of each video are also increased. 
The set was divided into training and testing sets in a class-balanced manner in a 4:1 ratio. 

For our analysis, we leveraged a pre-trained 3D-CNN model (Base Model) designed for binary classification of Congenital Heart Diseases in newborns. To enhance the model’s performance, we modified its architecture as specified in the main paper (\our{}$_\lambda$) and evaluated the results with varying parameter $\lambda$ values, ranging from 0.1 to 1. Our experiment aimed to assess the impact of these architectural modifications on the model's accuracy and early exit. The results presented in Tab.~\ref{tab:ultrasound} highlighted that \our{} improved the model's efficiency in terms of Normalized Exit Time (NET), while still maintaining competitive accuracy levels. While the base model consistently achieved the highest accuracy ($94.00\%$), \our{}$_{0.9}$ demonstrated a significant reduction in NET with only small accuracy drop ($-3.1\%$). This indicates that by adjusting $\lambda$, it is possible to balance nearly optimal accuracy with early exit decisions, allowing the model to make predictions faster without sacrificing performance.

\begin{figure*}[thb]
    % \vspace{-10pt}
    \centering
    \includegraphics[width=0.47\textwidth, trim={0 0.5cm 0 5.1cm},clip]{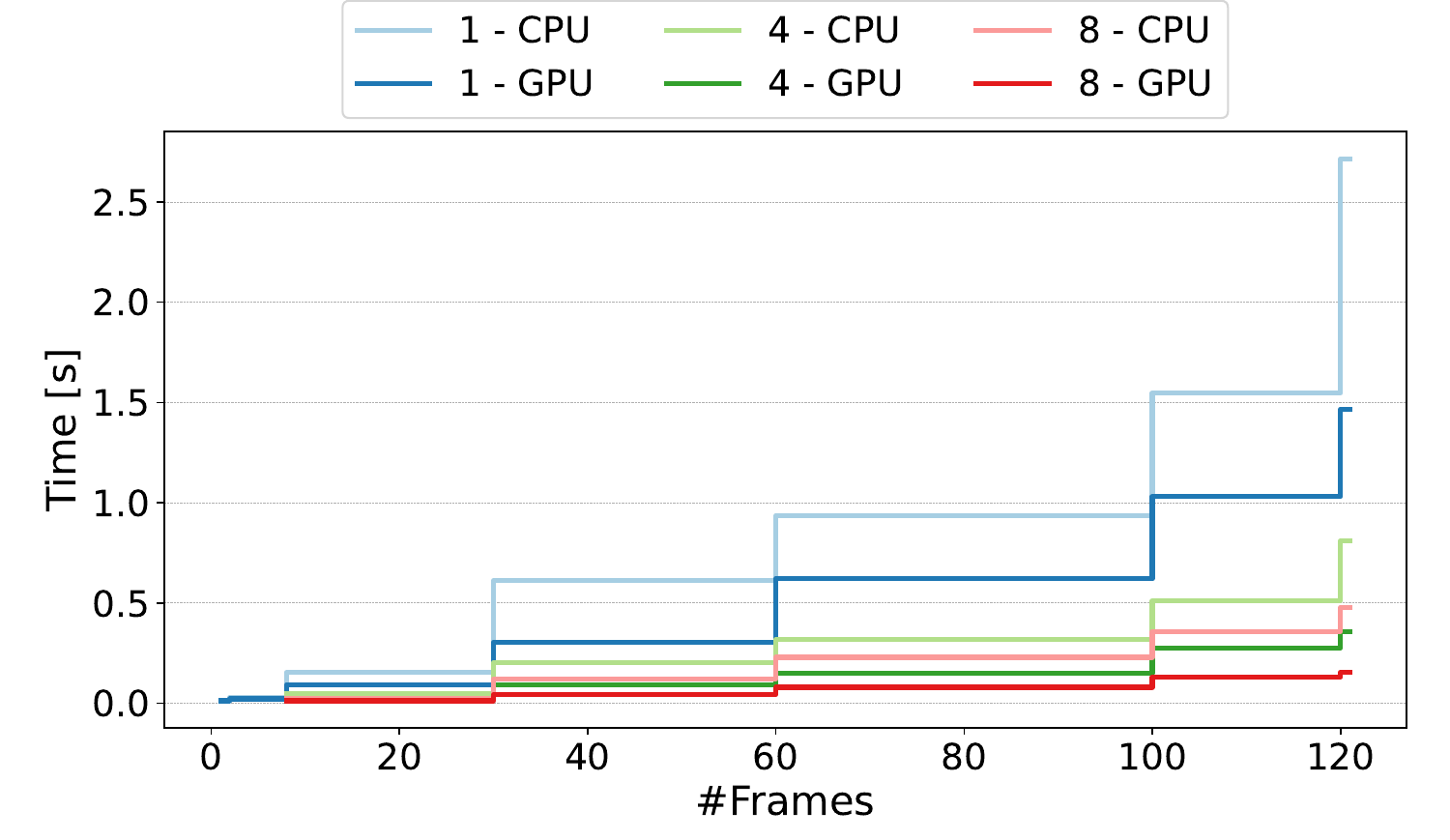}
    \caption{Model evaluation on time. Each line corresponds with the device used for evaluation and the number of frames inserted into the model in one step.}
    \vspace{-10pt}
    \label{fig:times}
    \vspace{15pt}
\end{figure*}

\paragraph{\our{} Performance Evaluation}

In this experiment the model was repeatedly evaluated on the selected video with different frame delivery parameters. The average time from 100 evaluations was taken as the result of the experiment for each set of parameters. The first parameter was the total number of frames delivered after all steps were completed. 
%To obtain a smaller number of frames, frames that did not fit were cut off. Of course, this resulted in a significant drop in accuracy, but only time was important in this experiment. 
Due to the growing history in memory, it was expected that the increase in time would not be linear, i.e., 2 times as many frames should result in an evaluation more than 2 times longer. The experimental results confirmed this thesis. On the other hand, not single frames but entire batches can be loaded into the model. For this reason, in the second experiment, the first experiment was repeated, but instead of a single frame, 4 or 8 frames were fed to the model in each step. Thus, the history stored in the model worked in the same way as before, but the number of reads and writes to it was reduced. Due to the expected operation of the model on devices with lower performance, e.g., mobile devices, the first part of the tests was carried out on a CPU. 
%due to the popularity of using AI models on GPU, t
The experiments were also repeated using the CUDA architecture. The results from both devices are presented in Fig.~\ref{fig:times}.

\begin{table*}[thb] 
\vspace{10pt}
\caption{Performance comparison of \our{}$_\lambda$ trained with different values of the parameter $\lambda$ and Base Model on the ultrasound Doppler dataset. The table reports classification accuracy ($\%$) and Normalized Exit Time (NET) for each model. The best results for each metric are \textbf{bolded}. The highest accuracy ($94.00\%$) is achieved by the Base Model, indicating that the default configuration outperforms \our{}$_\lambda$ versions. Among the \our{}$_\lambda$ models, $\lambda = 0.6$ yield the highest accuracy ($91.30\%$). The lowest NET value (0.2679), suggesting the earliest exit, is observed at $\lambda = 0.9$. Note that the \our{}$_\lambda$ does not show much difference in accuracy depending on the $\lambda$ parameter.}
\vspace{15pt}
\label{tab:ultrasound}
\centering
\resizebox{\linewidth}{!}{ % 
\begin{tabular}{cccccccccccc}
\toprule
$\boldsymbol{\lambda}$         & 0.1   & 0.2   & 0.3   & 0.4   & 0.5   & 0.6   & 0.7   & 0.8   & 0.9   & 1.0   & \textbf{Base Model} \\ \midrule
\textbf{Accuracy{(}\%{)}} & 86.95 & 88.53 & 88.53 & 90.90 & 90.90 & 91.30 & 91.30 & 90.90 & 90.90 & 88.93 & \textbf{94.00} \\ \midrule
\textbf{NET} & 0.3410 & 0.3869 & 0.3316 & 0.4475 & 0.3454 & 0.3773 & 0.3058 & 0.3557 & \textbf{0.2679} & 0.3581 & 1 \\ \bottomrule
\end{tabular}
} % 
\end{table*}

\end{document}